\documentclass{article} 
\usepackage[preprint]{tmlr}

\usepackage{amsmath,amsfonts,amssymb}
\usepackage{booktabs}
\usepackage{graphicx}
\usepackage{microtype}
\usepackage{xspace}
\usepackage{hyperref}
\usepackage{url}

\newcommand{\keydiff}{KeyDiff\xspace}
\newcommand{\snapkv}{SnapKV\xspace}
\newcommand{\hho}{H2O\xspace}
\newcommand{\sllm}{StreamingLLM\xspace}
\newcommand{\besttriv}{best-of-3 trivial baseline\xspace}
\newcommand{\ci}[2]{[#1,\,#2]}

\title{How Query Visibility Changes\\KV-Cache Compression Rankings:\\A Matched-Budget Audit}

\author{%
\name Daming Luo \email daming.luo@student.uts.edu.au \\
\addr University of Technology Sydney
\AND
\name Christy Liang \email Jie.Liang@uts.edu.au \\
\addr University of Technology Sydney
\AND
\name Junyu Xuan \email Junyu.Xuan@uts.edu.au \\
\addr University of Technology Sydney}

\def\month{07}
\def\year{2026}
\def\openreview{\url{https://openreview.net/forum?id=XXXX}}

\begin{document}

\maketitle

\begin{abstract}
KV-cache compression methods are predominantly evaluated with the query
appended to the context \emph{before} compression---a \textbf{query-aware}
protocol. Yet the economic case for a compressed KV cache is \emph{reuse}:
compress a document once, answer many future questions against it. In that
deployment, compression must happen \textbf{query-agnostic}---before any
question is seen. We present a matched-budget audit of six published
compression methods against three trivial baselines on three open 7--9B
models (144{,}300 paired evaluations on RULER-8192; 40{,}800 on LongBench;
50{,}000-resample paired bootstrap throughout). Everything is held
fixed---model, compression ratio, instances, decoding---except the scoring
rule. Three findings. \textbf{(1)}~Query visibility changes the rankings:
under the agnostic protocol, of the five audited methods that share a
common attention backend, only \keydiff beats a \besttriv consistently
(31/36 cells), and the most widely deployed method, \snapkv, \emph{loses} to
``keep the start and the recent window'' on average ($-0.066$).
\textbf{(2)}~The per-method drop between the two protocols is
\emph{ordered} consistently with how visible the question is to each
method's scoring signal, legible in its source code: from
$\Delta{=}{+}0.198$ for \snapkv (the question sits inside its 64-token
observation window) down to $\Delta{=}{+}0.011$ for \keydiff (its score
contains no query term at all). We advance this as a mechanistic
hypothesis---ordinal evidence over six methods, not a fitted law. Under
that reading, query-aware scores partly measure \emph{query relevance}
rather than the \emph{information importance} that cache reuse requires.
\textbf{(3)}~The audit surfaced two reproducible methodological hazards:
an \emph{attention-backend confound}---swapping
\texttt{sdpa} for \texttt{eager} on an uncompressed model shifts RULER
accuracy by $-0.29$, larger than most method-vs-baseline gaps, which forces
us to withdraw any ranking claim about the one method (\hho) that requires
the eager backend---and a \emph{tokenizer-dependent benchmark length}:
RULER's nominal ``8192'' overflows gemma-2's positional budget by up to
30\%, silently zeroing 7 of 13 subtasks even without compression. We release
the audit harness, all per-instance records, and the paired statistics.
\end{abstract}

\section{Introduction}
\label{sec:intro}

Long-context inference is memory-bound: the KV cache of an 8B model at 128k
tokens exceeds the weights themselves. A large literature therefore prunes
the cache---score every cached token, evict the bottom fraction---and reports
that 50--90\% of the cache can be dropped with little accuracy loss.

Most of that evidence is collected under one quiet convenience: the
benchmark question is placed in (or after) the context before compression
runs, so the compressor's scoring pass \emph{sees the question}. We call
this the \textbf{query-aware} protocol. It matches a deployment in which
every question re-reads and re-compresses the document---a deployment in
which the KV cache buys almost nothing, because the dominant cost being
amortised (prefill) is paid again per question. The setting that makes a
compressed cache economically interesting is the opposite one:
\textbf{compress once, query many times}---a contract interrogated by dozens
of questions, a codebase queried all day. There, compression is necessarily
\textbf{query-agnostic}: the question does not exist yet when eviction
happens.

The concern itself is not new. SCBench \citep{li2025scbench} documents,
across a KV-cache lifecycle benchmark, that query-dependent long-context
methods degrade when the query is unavailable, and a recent line of methods
is explicitly designed query-agnostic \citep{kim2025kvzip,
chari2025compactor, devoto2025expected}. What the literature does not yet
provide is a \emph{measurement}: how large each published method's
dependence on query visibility is, isolated from every other variable. This
paper asks that narrow, auditable question:
\begin{quote}
\emph{Under a matched budget, how much of each published method's reported
gain survives the move from query-aware to query-agnostic compression---and
does the size of the drop track how the method's scoring signal uses the
question?}
\end{quote}

We answer it with an audit, not a new method. Six published presses
(\snapkv, \hho, TOVA, ExpectedAttention, AdaKV, \keydiff) are run against
three \emph{trivial} baselines (random eviction, key-norm, \sllm's ``start +
recent window'') and a full-cache anchor, with model, ratio, instances and
decoding held identical; only the press varies. Both protocols are run over
the full grid, so the agnostic$\to$aware delta is a \emph{within-model,
within-instance} paired contrast---the cleanest contrast the question
admits.

\paragraph{Contributions.}
\begin{enumerate}
\item \textbf{A matched-budget audit protocol} with trivial-baseline
  anchoring, exact per-instance pairing, enforced coverage checks, and a
  best-of-selection debias---plus two hazards we hit that any such audit
  must control: attention-backend mixing (\S\ref{sec:backend}) and dedup
  keys that omit the backend (\S\ref{sec:stats}).
\item \textbf{A quantitative measurement of query dependence}: per-method
  agnostic$\to$aware deltas across three models under one unified protocol
  (\S\ref{sec:headline}), together with a mechanistic reading---the deltas
  are ordered consistently with the visibility of the question in each
  method's \texttt{score()} implementation---which we state as a
  hypothesis supported by ordinal evidence over six methods, not as a
  fitted law (\S\ref{sec:mechanism}).
\item \textbf{The deployment-protocol picture}: under query-agnostic
  compression on RULER, of the five backend-comparable methods only the
  query-independent one (\keydiff) beats the \besttriv (31/36 cells,
  $+0.171$), while \snapkv averages below it (13/36, $-0.066$); a LongBench
  check shows the picture survives on natural text in \emph{validity} but
  not in \emph{exclusivity}---other methods catch up or overtake
  (\S\ref{sec:results}--\ref{sec:longbench}).
\item \textbf{Two reproducible evaluation hazards}: an attention-backend
  confound (eager vs.\ sdpa shifts pooled RULER accuracy by $-0.221$,
  larger than most method-vs-baseline gaps; it removes \hho from our
  ranking conclusions) and tokenizer-dependent benchmark length (RULER's
  nominal 8192 overflows gemma-2's positional budget by up to 30\%,
  silently zeroing 7 of 13 subtasks with no compression at all)
  (\S\ref{sec:backend}--\ref{sec:tokenizer}).
\end{enumerate}

We emphasise what this paper is not. It is not a new compression method, not
an endorsement of \keydiff, and not a claim that query-aware papers are
wrong on their own terms---we \emph{reproduce} \snapkv's query-aware gain
($+0.132$). Nor is it the first observation that query visibility matters:
\citet{li2025scbench} make that point qualitatively. It is the controlled,
per-method quantification of the effect, and a documentation of what such
an audit must hold fixed to be valid.

\section{Background and related work}
\label{sec:related}

\paragraph{KV-cache eviction.}
\hho \citep{zhang2023h2o} keeps ``heavy hitters'' by accumulated attention
mass; \snapkv \citep{li2024snapkv} scores tokens by attention from the last
64 positions (an ``observation window''); TOVA \citep{oren2024tova} uses the
final token's attention; AdaKV \citep{feng2025adakv} reallocates a
\snapkv-style budget per-head; ExpectedAttention \citep{devoto2025expected}
scores by an analytic estimate of future query attention; \keydiff
\citep{park2025keydiff} departs from attention entirely and keeps tokens
whose \emph{key vectors} are angular outliers from the mean key direction.
\sllm \citep{xiao2024streamingllm}---retain sink + recent window---is here
demoted to a \emph{trivial baseline}, alongside random eviction and the
key-norm heuristic of \citet{devoto2024knorm}, which retains
\emph{low}-norm keys (low key norm correlates with high attention). All
presses as implemented in NVIDIA kvpress 0.5.4 \citep{kvpress}.

\paragraph{Shared-context evaluation and SCBench.}
The closest prior work is SCBench \citep{li2025scbench}, which evaluates
long-context methods across a full KV-cache lifecycle (multi-turn and
multi-request modes) and reports qualitatively that query-dependent
compression---\snapkv{} in particular---struggles when the query is not
visible at compression time. Our audit is narrower and, on that single
point, sharper: we hold model, budget, instances and decoding fixed, pair
every record, anchor against trivial baselines, and measure the per-method
agnostic$\to$aware delta with bootstrap confidence intervals---turning the
qualitative warning into a quantity attributable to each scoring rule. The
two efforts are complementary: SCBench varies the workload; we isolate the
protocol variable.

\paragraph{Query-agnostic compression methods.}
A recent line of work designs for the reuse setting directly: KVzip
\citep{kim2025kvzip} scores KV pairs by their contribution to
reconstructing the context, Compactor \citep{chari2025compactor} uses
query-agnostic approximate leverage scores, and ExpectedAttention
\citep{devoto2025expected} integrates over a modelled future-query
distribution. Of these, ExpectedAttention is in our audit; KVzip's
multi-pass compression procedure does not fit a single-pass matched-budget
harness, and Compactor is not implemented in the audited kvpress release
(both are natural extensions, \S\ref{sec:limitations}). Our results support
this line's motivation while adding a caution it inherits: on our grid the
analytically query-agnostic ExpectedAttention still loses to a
keep-start-plus-recent baseline (\S\ref{sec:headline}), so query-agnostic
\emph{design} does not by itself guarantee agnostic-protocol
\emph{wins}---which is what an audit is for.

\section{The audit}
\label{sec:audit}

\subsection{Design principle: one variable}
A cell of the audit fixes (model, benchmark axis, compression ratio,
protocol arm) and varies only the press. Within a cell every press answers
\emph{the same 650 instances} with \emph{the same decoding} (greedy, fixed
\texttt{max\_new\_tokens}) at \emph{the same cache budget} (uniform
compression ratio $r \in \{0.25, 0.5, 0.75, 0.9\}$). A \textsc{FullCache}
anchor ($r{=}0$) calibrates each cell's headroom.

\subsection{Presses and baselines}
Six \textbf{real} presses: \snapkv, \hho (ObservedAttention), TOVA,
ExpectedAttention, AdaKV, \keydiff. Three \textbf{trivial} presses: Random,
Knorm (keep lowest-norm keys), \sllm (keep first $n_{\text{sink}}$ + recent
window). The comparison target throughout is \textbf{best-of-3 trivial}:
the max of the three trivial scores \emph{per cell}. This is deliberately
adversarial to the real methods---a method that cannot beat the best of
three near-zero-cost rules has no deployment case---and we correct the
residual best-of-selection bias in \S\ref{sec:debias}. Empirically the best
trivial is \sllm in 27/36 agnostic cells, Knorm in 8, Random in 1: ``beats
best-of-3 trivial'' in practice means ``beats keep-start-plus-recent''.

\subsection{Benchmarks, models, protocols}
\textbf{RULER-8192} \citep{hsieh2024ruler}: 13 subtasks, 650 instances,
grouped into three axes---retrieval (needle variants), aggregation
(CWE/FWE), multi-hop (VT/QA). \textbf{LongBench} \citep{bai2024longbench}
(16 English tasks $\times$ 50 instances) serves as the natural-text
robustness check (\S\ref{sec:longbench}). Models: Llama-3.1-8B-Instruct
(Meta), Qwen2.5-7B-Instruct (Alibaba), DeepSeek-R1-Distill-Qwen-7B
(reasoning-tuned; shares Qwen lineage---it contributes statistical, not
architectural, diversity, which we state as a limitation). A fourth lineage
(gemma-2-9b) was disqualified by the benchmark itself
(\S\ref{sec:tokenizer}).

\paragraph{Protocol arms.}
\emph{Query-aware}: the question is visible to the press's scoring pass.
\emph{Query-agnostic}: the context is compressed first; the question is
appended after eviction. Both arms share everything else, so per-instance
differences within (model, press, ratio) isolate the protocol.

\paragraph{Scale.}
RULER grid: 3 models $\times$ 2 arms $\times$ (1 anchor + 9 presses
$\times$ 4 ratios) $\times$ 650 instances = \textbf{144{,}300 records, zero
errors, zero holes} (coverage enforced programmatically; 331 OOM holes
detected and refilled during the run). LongBench adds 40{,}800 records; the
backend control (\S\ref{sec:backend}) adds 600.

\subsection{Statistics---and two pitfalls}
\label{sec:stats}
All contrasts are \textbf{paired per instance} ($B{=}50{,}000$ bootstrap
resamples); cross-model ``rank flip'' probabilities use joint
instance-aligned resampling. Two operational pitfalls we hit deserve
documentation because they silently corrupt audits of this shape.
\emph{(i)~Dedup keys must include the attention backend.} Ours was (model,
press, ratio, task, instance); when \hho---which requires \texttt{eager}
attention---was backfilled, trivial-baseline cells were skipped as
``already done'', leaving zero non-\hho eager rows and an unmeasured
confound we later had to control explicitly (\S\ref{sec:backend}).
\emph{(ii)~Coverage must be asserted, not assumed}: our checker compares
the exact instance set per press per cell and is what surfaced the 331 OOM
holes.

\section{Results on RULER}
\label{sec:results}

\subsection{Anchors}
\textsc{FullCache} scores (agnostic arm) delimit each model's headroom
(Table~\ref{tab:anchors}). R1-Distill's low anchors (it spends its budget
reasoning, not retrieving) produce floor effects that matter in
\S\ref{sec:flips}.

\begin{table}[t]
\caption{\textsc{FullCache} anchor scores per model and axis (agnostic arm,
$r{=}0$). These delimit the headroom available to any press in the cells
below them.}
\label{tab:anchors}
\begin{center}
\begin{tabular}{lccc}
\toprule
\textbf{Model} & \textbf{Retrieval} & \textbf{Aggregation} & \textbf{Multi-hop} \\
\midrule
Llama-3.1-8B        & 1.000 & 0.964 & 0.825 \\
Qwen2.5-7B          & 0.993 & 0.948 & 0.753 \\
R1-Distill-Qwen-7B  & 0.667 & 0.624 & 0.285 \\
\bottomrule
\end{tabular}
\end{center}
\end{table}

\subsection{Headline: the protocol decides who wins}
\label{sec:headline}

\begin{figure}[t]
\begin{center}
\includegraphics[width=\textwidth]{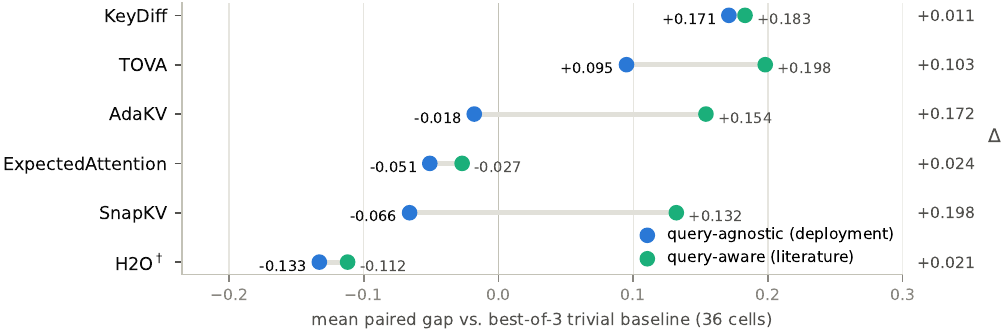}
\end{center}
\caption{\textbf{The protocol flip.} Each press's mean paired gap vs the
\besttriv over all 36 cells, under the deployment protocol (query-agnostic,
blue) and the literature's protocol (query-aware, green); the right margin
gives the per-press protocol effect $\Delta$. \keydiff barely moves; \snapkv
and AdaKV swing from losing to winning on the strength of seeing the
question. $^\dagger$backend-confounded (\S\ref{sec:backend}).}
\label{fig:flip}
\end{figure}

Table~\ref{tab:headline} and Figure~\ref{fig:flip} give the per-press
verdict vs the \besttriv over the 36 cells (3 models $\times$ 3 axes
$\times$ 4 ratios); ``wins'' = positive mean paired gap in the cell.

\begin{table}[t]
\caption{Per-press verdict vs the \besttriv over 36 cells. $\Delta$ is the
protocol effect (aware $-$ agnostic mean gap). $^\dagger$\hho runs on the
\texttt{eager} backend while every other row runs on \texttt{sdpa};
\S\ref{sec:backend} shows the backend alone shifts scores by more than
\hho's deficit, so we make \emph{no ranking claim} about \hho---the row is
reported for completeness only.}
\label{tab:headline}
\begin{center}
\begin{tabular}{lccccc}
\toprule
& \multicolumn{2}{c}{\textbf{Query-agnostic}} & \multicolumn{2}{c}{\textbf{Query-aware}} & \\
\cmidrule(lr){2-3}\cmidrule(lr){4-5}
\textbf{Press} & wins & mean gap & wins & mean gap & $\boldsymbol{\Delta}$ \\
\midrule
\keydiff            & \textbf{31/36} & $\boldsymbol{+0.171}$ & 33/36 & $+0.183$ & $\boldsymbol{+0.011}$ \\
TOVA                & 21/36 & $+0.095$ & 33/36 & $+0.198$ & $+0.103$ \\
AdaKV               & 16/36 & $-0.018$ & 28/36 & $+0.154$ & $+0.172$ \\
ExpectedAttention   & 14/36 & $-0.051$ & 15/36 & $-0.027$ & $+0.024$ \\
\snapkv             & 13/36 & $-0.066$ & 25/36 & $+0.132$ & $\boldsymbol{+0.198}$ \\
\midrule
\hho$^\dagger$      &  4/36 & $-0.133$ &  9/36 & $-0.112$ & $+0.021$ \\
\bottomrule
\end{tabular}
\end{center}
\end{table}

Read column by column: under the \emph{aware} protocol four of the five
backend-comparable methods beat the trivial baselines---the literature's
picture, which we reproduce (\snapkv $+0.132$). Under the \emph{agnostic}
protocol the picture inverts: \textbf{among the audited methods only
\keydiff survives}, and \snapkv---the most widely adopted method in this
family---averages \emph{below} ``keep the start and the recent window''.

\subsection{Where the wins live}
\label{sec:axes}

\begin{table}[t]
\caption{Per-axis mean gap vs the \besttriv (12 cells per entry: 3 models
$\times$ 4 ratios), agnostic arm. Parentheses give cells won.}
\label{tab:axes}
\begin{center}
\small
\begin{tabular}{lccc}
\toprule
\textbf{Press} & \textbf{Retrieval} & \textbf{Aggregation} & \textbf{Multi-hop} \\
\midrule
\keydiff          & $\boldsymbol{+0.249}$ (12/12) & $\boldsymbol{+0.201}$ (11/12) & $+0.064$ (8/12) \\
TOVA              & $+0.115$ (7/12)  & $+0.124$ (8/12)  & $+0.047$ (6/12) \\
AdaKV             & $-0.021$ (4/12)  & $-0.042$ (5/12)  & $+0.009$ (7/12) \\
ExpectedAttention & $-0.067$ (4/12)  & $-0.116$ (2/12)  & $+0.030$ (8/12) \\
\snapkv           & $-0.092$ (3/12)  & $-0.088$ (3/12)  & $-0.018$ (7/12) \\
\midrule
\hho$^\dagger$    & $-0.232$ (0/12)  & $-0.129$ (2/12)  & $-0.037$ (2/12) \\
\bottomrule
\end{tabular}
\end{center}
\end{table}

\begin{figure}[t]
\begin{center}
\includegraphics[width=\textwidth]{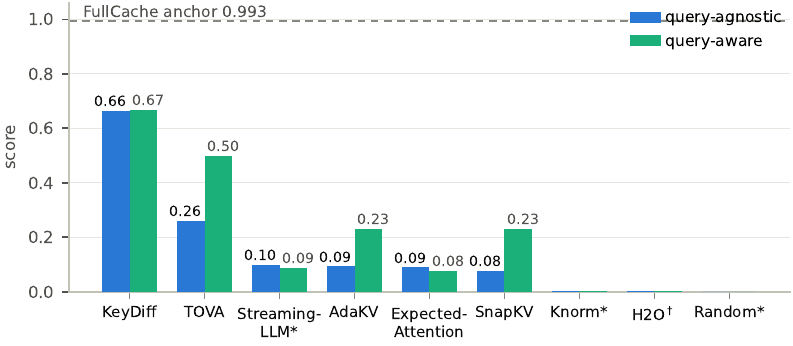}
\end{center}
\caption{\textbf{The single sharpest cell} (Qwen2.5-7B / retrieval /
$r{=}0.9$). With 90\% of the cache evicted on a task the uncompressed model
solves (anchor 0.993, dashed), \snapkv and AdaKV triple their score when the
question is visible during compression; \keydiff's score does not depend on
it. *trivial baseline; $^\dagger$backend-confounded
(\S\ref{sec:backend}).}
\label{fig:cell}
\end{figure}

\keydiff's advantage concentrates exactly where compression is hardest and
the stakes highest---retrieval at high ratios (Table~\ref{tab:axes}). The
single sharpest cell (Qwen2.5-7B, retrieval, $r{=}0.9$, anchor 0.993) is
shown in Figure~\ref{fig:cell}: \keydiff 0.663, TOVA 0.260, \sllm 0.100,
AdaKV 0.093, ExpectedAttention 0.090, \snapkv 0.077, Knorm 0.003, \hho
0.003, Random 0.000. The same cell under the aware protocol: \snapkv
recovers to 0.230 and TOVA to 0.500---seeing the question is worth
$3\times$ to \snapkv.

\keydiff is not ``secretly not compressing'': on Llama retrieval its score
falls monotonically $0.940 \to 0.817 \to 0.727 \to 0.677$ across
$r = 0.25 \to 0.9$. It pays for compression; it just evicts better.

\subsection{Group-level verdict, leave-one-out, and selection debias}
\label{sec:debias}

Best-of-6-real vs best-of-3-trivial: real wins 34/36 (agnostic). But this
comparison is load-bearing on one method: \textbf{leave-one-out removes any
single press and real still wins 34/36---except \keydiff, whose removal
collapses the verdict to 25/36.} In the aware arm, removing \emph{any}
press leaves 34/36: four methods can carry it. The agnostic-arm ``real
methods win'' story is a one-method story.

Best-of-6 vs best-of-3 is also structurally biased toward the larger
roster. Debiasing by expectation over all $\binom{6}{3}=20$ matched
best-of-3 subsets: agnostic $34 \to \textbf{29}/36$; aware $34 \to 34/36$.
The conclusion is unchanged but honest: the agnostic-arm margin is thinner
than the raw best-of comparison suggests.

\subsection{Cross-model consistency (and a headline we retracted)}
\label{sec:flips}

An earlier snapshot of this audit (3--5 real presses, 331 unfilled OOM
holes) showed cross-model \emph{rank flips} with probability 1.000 on two
axes---models disagreeing on whether real methods beat trivial ones.
Completing the roster and the holes killed that finding: flip probabilities
on retrieval and aggregation drop to 0.000--0.252, and the four remaining
flip cells are all contributed by R1-Distill in cells where its gap is
within noise of zero (e.g.\ $-0.007$)---a floor effect (its multi-hop
anchor is 0.285), not an architecture effect. We report this retraction
because an audit framework that cannot kill its author's favourite finding
is not an audit.

\section{A mechanistic hypothesis: the delta tracks query visibility}
\label{sec:mechanism}

\begin{figure}[t]
\begin{center}
\includegraphics[width=\textwidth]{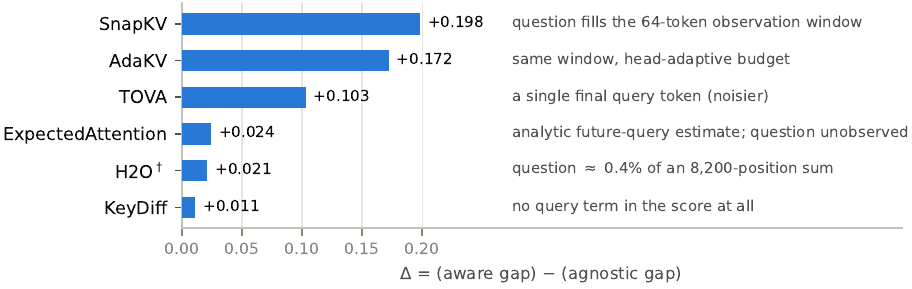}
\end{center}
\caption{\textbf{The agnostic$\to$aware delta per method, ordered.} The
ordering is consistent with the weight the question carries in each
method's scoring signal, read from the \texttt{score()} implementations;
we treat this correspondence as a mechanistic hypothesis
(\S\ref{sec:mechanism}).}
\label{fig:mech}
\end{figure}

Why does the protocol move \snapkv by $+0.198$ but \keydiff by $+0.011$?
A candidate answer is legible in each press's scoring function (kvpress
0.5.4): the measured deltas are ordered consistently with how visible the
question is to each scoring signal (Table~\ref{tab:mechanism},
Figure~\ref{fig:mech}). We are explicit about the strength of this
evidence: six methods give six correctly ordered points---an
\emph{empirical ordering}, not a fitted quantitative law. We advance it as
a mechanistic hypothesis; the manipulations that would test it directly
(varying observation-window content continuously, injecting mismatched
questions at compression time) are outside this audit's scope.

\begin{table}[t]
\caption{Each press's scoring signal, the visibility of the question inside
it, and the measured protocol effect $\Delta$. The measured ordering of
$\Delta$ is consistent with the question's visibility, read from source;
with six methods this is ordinal evidence for a hypothesis, not a fitted
law.}
\label{tab:mechanism}
\begin{center}
\small
\begin{tabular}{lllc}
\toprule
\textbf{Press} & \textbf{Scoring signal (source)} & \textbf{Question's visibility} & $\boldsymbol{\Delta}$ \\
\midrule
\snapkv & attention from last 64 tokens (pooled) & fills the window & $\boldsymbol{+0.198}$ \\
AdaKV & \snapkv's scorer, head-adaptive & same window & $+0.172$ \\
TOVA & attention from the last token & 1 noisy query token & $+0.103$ \\
ExpectedAttention & analytic future-query estimate & not directly observed & $+0.024$ \\
\hho & attention summed over all ${\sim}8{,}200$ positions & ${\approx}0.4\%$ of the sum & $+0.021$ \\
\keydiff & $-\cos(k_i, \bar{k})$; \textbf{no query term} & zero & $\boldsymbol{+0.011}$ \\
\bottomrule
\end{tabular}
\end{center}
\end{table}

The hypothesis: $\Delta$ grows with \textbf{the fraction of the scoring
signal contributed by the question.} Under this reading, methods whose
observation window is ``the end of the context'' are not measuring which
tokens are important; they are measuring which tokens are relevant \emph{to
whatever happens to sit at the end}---in query-aware evaluation, the answer
key itself. Remove the answer
key (agnostic arm) and the window contains the tail of the document, which
has no privileged relation to a question that arrives later.

\keydiff survives \emph{because it never asks}. Its criterion---keep tokens
whose key vectors point away from the bulk---is a query-independent notion
of informativeness: repetitive filler produces tightly clustered keys;
content that is unusual produces outliers. Whether that is the \emph{right}
notion is a benchmark-dependent question (\S\ref{sec:longbench}); that it
is \emph{protocol-immune} is a theorem of its formula.

We conjectured (and the attention-sink literature supports) a mechanism for
\hho's weakness---accumulated attention concentrates on early sink tokens,
so \hho approximately recomputes \sllm's hard-coded rule at eager-attention
prices---but per \S\ref{sec:backend} we can no longer present \hho's
measured rank as evidence for it.

\section{Robustness checks and negative results}
\label{sec:robustness}

\subsection{Natural text: \keydiff keeps its validity, loses its monopoly}
\label{sec:longbench}

RULER's haystack is pathologically repetitive, which maximises exactly the
key-outlierness \keydiff scores. Prediction registered before the
experiment: \keydiff's edge should shrink on natural prose. On LongBench
(16 tasks, both ratios, agnostic), \keydiff vs the \besttriv:

\begin{table}[t]
\caption{LongBench (natural text, agnostic): \keydiff's mean gap vs the
\besttriv.}
\label{tab:longbench}
\begin{center}
\begin{tabular}{lcc}
\toprule
\textbf{Model} & $\boldsymbol{r=0.5}$ & $\boldsymbol{r=0.75}$ \\
\midrule
Llama-3.1-8B & $\boldsymbol{+0.019}$ & $\boldsymbol{+0.055}$ \\
Qwen2.5-7B   & $\boldsymbol{+0.072}$ & $\boldsymbol{+0.088}$ \\
R1-Distill   & $-0.011$ & $-0.017$ \\
\bottomrule
\end{tabular}
\end{center}
\end{table}

\begin{figure}[t]
\begin{center}
\includegraphics[width=\textwidth]{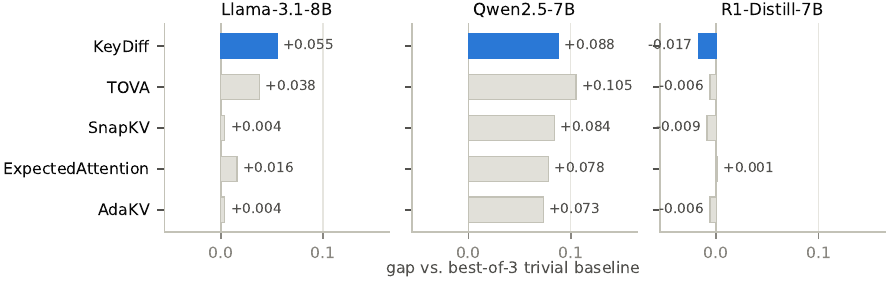}
\end{center}
\caption{\textbf{On natural text \keydiff stays ahead of the trivial
baselines on the two healthy-anchor models---but is no longer alone}
(agnostic, $r{=}0.75$ shown). Unlike on RULER, the other real presses
(gray) catch up or overtake.}
\label{fig:longbench}
\end{figure}

Its \emph{validity} survives: positive on both healthy-anchor models (4/6
cells; R1's LongBench scores sit near floor throughout,
Table~\ref{tab:longbench}). Its \emph{exclusivity} does not: on natural
text other real presses catch up or pass it (Qwen $r{=}0.75$: TOVA $+0.105
>$ \keydiff $+0.088$; Qwen $r{=}0.5$: ExpectedAttention $+0.084 >$ \keydiff
$+0.072$; Figure~\ref{fig:longbench}). RULER inflated \keydiff's
uniqueness, not its effectiveness. Our claim is therefore the weaker,
better-supported one: \textbf{\keydiff is the only audited method that is
query-independent and effective on both benchmark families}---not ``the
best method''.

LongBench is run in the agnostic arm only, by design: the question it
exists to answer---whether \keydiff's agnostic-arm win is an artefact of
RULER's synthetic haystack---lives entirely in that arm, while the
aware-arm picture is established on the full RULER grid
(\S\ref{sec:headline}).

\subsection{The attention-backend confound: a ranking withdrawn}
\label{sec:backend}

\hho needs explicit attention weights, hence the \texttt{eager} backend;
everything else ran \texttt{sdpa}. To bound the confound we re-ran
\textsc{FullCache}, \keydiff and \sllm under \texttt{eager} on Qwen2.5-7B
(agnostic; $r \in \{0.25, 0.9\}$; 10 subtasks; 600 records exactly paired
to their sdpa twins):
\begin{quote}
pooled $\;\text{score}(\texttt{eager}) - \text{score}(\texttt{sdpa}) =
\mathbf{-0.221}$, \quad 95\% CI $\ci{-0.255}{-0.188}$.
\end{quote}

\begin{figure}[t]
\begin{center}
\includegraphics[width=\textwidth]{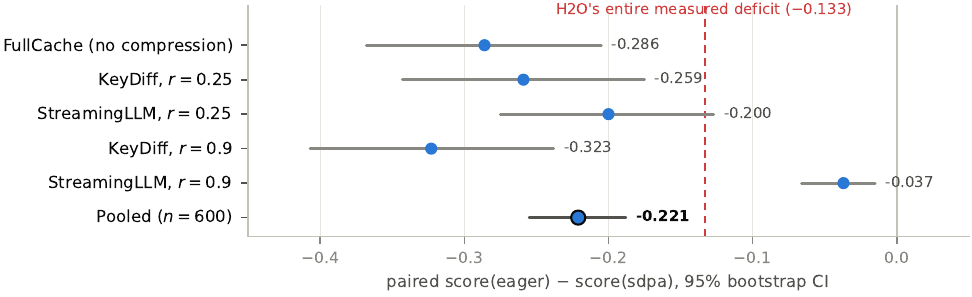}
\end{center}
\caption{\textbf{The backend control.} Paired eager$-$sdpa deltas with 95\%
bootstrap CIs; every row is the same model, instances and budget with only
the attention kernel changed. The pooled effect---and even the uncompressed
\textsc{FullCache} row---exceeds \hho's entire deficit (red dashed).}
\label{fig:backend}
\end{figure}

The CI excludes zero and its magnitude \emph{exceeds} \hho's entire
measured deficit ($-0.133$). Most striking, the \textbf{uncompressed}
\textsc{FullCache} row falls $0.897 \to 0.611$ ($-0.286$)---on a model with
no logit softcap, where the two backends are mathematically equivalent. The
divergence is numerical, not semantic: fp16 accumulation-order differences
flip an early greedy argmax and the retrieval transcript derails. Its sign
does not mean ``eager is worse''; it means the two backends \emph{diverge
chaotically} at 8k context under greedy decoding.

Consequences. \emph{(i)}~\textbf{We withdraw the ranking claim for \hho}:
its deficit is observationally inseparable from a backend term larger than
itself. \emph{(ii)}~The within-sdpa comparisons---everything else in this
paper---are unaffected: the backend is held fixed there by construction.
\emph{(iii)}~The general lesson is the paper's thesis in miniature:
\emph{matched-budget means matching everything, including the attention
kernel}. Any audit that mixes backends across arms carries a hidden
${\sim}0.2$ confound, and nothing in standard logs will disclose it. The
control is measured on one model and arm; the three longest subtasks are
excluded because eager's $O(L^2)$ fp32 softmax exhausts 48\,GB at
$L\approx11$k (all three axes remain covered), and we state both scope
limits explicitly.

\subsection{The tokenizer-length trap: a model disqualified by the benchmark}
\label{sec:tokenizer}

We attempted a fourth lineage (gemma-2-9b,
\texttt{max\_position\_embeddings} = 8192). RULER's nominal ``8192'' is
counted in a reference tokenizer; gemma's tokenizer renders the same
contexts as up to ${\sim}10{,}600$ tokens, overrunning its positional
budget on 7 of 13 subtasks (350/650 instances, including all three
\texttt{niah\_single}). On those subtasks gemma's \emph{uncompressed}
\textsc{FullCache} scores 0.000 while scoring 0.907--0.925 on the six that
fit. This is not a harness bug---the model cannot hold the benchmark---and
no standard log line reports it. \textbf{Observation: benchmark nominal
lengths are tokenizer-dependent; any RULER-8192 result for gemma-2 in the
literature was produced under silent RoPE extrapolation.} (A separate, real
transformers bug---\texttt{sdpa} silently dropping gemma's
\texttt{attn\_logit\_softcapping}---was verified but is \emph{not} the
cause; the eager backend reproduces the zeros. All 4{,}932 invalid rows
were quarantined, not deleted.)

\section{Limitations}
\label{sec:limitations}

\begin{enumerate}
\item \textbf{Two benchmark families, one context length.} RULER-8192 +
  LongBench; real pressure lives at 32k--128k, where the ``recent window''
  fraction shrinks and we expect (but have not measured) the trap to
  sharpen.
\item \textbf{Architectural diversity is thin}: R1-Distill shares Qwen
  lineage; the Google lineage was disqualified (\S\ref{sec:tokenizer}). The
  headline is protocol-internal (each model vs itself), which does not
  require cross-model diversity, but the ranking claims would benefit from
  it.
\item \textbf{Six methods, kvpress implementations.} KVzip
  \citep{kim2025kvzip} (multi-pass compression, incompatible with the
  single-pass harness), Compactor \citep{chari2025compactor} (not in
  kvpress 0.5.4) and PyramidKV (upstream tensor bug at 0.5.4) are absent,
  and conclusions are about the audited implementations. Extending the
  audit to purpose-built query-agnostic methods is the most direct
  follow-up.
\item \textbf{\hho's rank is unresolved} (\S\ref{sec:backend})---
  deliberately, because resolving it requires an all-eager re-audit that
  the backend divergence itself shows would measure the backend as much as
  the method.
\item \textbf{Uniform-ratio budgets.} Head- or layer-adaptive budget
  \emph{allocation} under an agnostic protocol is the natural next
  question (we detect the trap; we do not yet detect, at inference time,
  which allocation a given input needs).
\item \textbf{The mechanism claim is ordinal.} Six methods give six
  correctly ordered points; establishing the visibility hypothesis
  quantitatively requires a continuous query-sensitivity measure (e.g.\
  counterfactual-query selection overlap, graded query exposure at
  compression time), which we leave to future work.
\end{enumerate}

\section{Conclusion}
\label{sec:conclusion}

Under a matched budget, a substantial part of what the audited methods'
query-aware scores measure is access to the question---an input that cache
reuse, the very deployment that motivates compression, cannot provide. The
dependence does not look incidental: across the six audited methods its
size is ordered by the question's visibility in each scoring signal,
legible in source code, and the one method with no query term is the one
whose performance the protocol cannot touch---an ordering we advance as a
mechanistic hypothesis awaiting direct manipulation. An audit built to be adversarial to its own
conclusions---trivial baselines, paired statistics, leave-one-out,
selection debias, backend controls---overturned its author's first
headline, withdrew a ranking, and disqualified a model; what survived is
correspondingly harder to explain away. We release the harness, the
185{,}700 paired records, and the statistics pipeline.

\subsubsection*{Broader impact statement}
This work audits evaluation practice; its main societal effect is to make
deployment decisions about KV-cache compression better calibrated, reducing
wasted compute and overclaimed capability. We identify no direct negative
applications.

\bibliography{refs}
\bibliographystyle{tmlr}

\appendix

\section{Reproducibility}
\label{app:repro}

\begin{itemize}
\item \textbf{Grid}: 3 models $\times$ 2 arms $\times$ (\textsc{FullCache}
  + 9 presses $\times \{0.25, 0.5, 0.75, 0.9\}$) $\times$ 650 RULER
  instances = 144{,}300 records; LongBench $16 \times 50 \times$ 8 presses
  (+anchor) $\times \{0.5, 0.75\} \times$ 3 models = 40{,}800; backend
  control 600.
\item \textbf{Decoding}: greedy; per-task \texttt{max\_new\_tokens};
  R1-Distill's \texttt{<think>} block force-closed (otherwise it burns the
  token budget and never answers---scoring all methods to zero uniformly).
\item \textbf{Scoring}: QA tasks---any-gold-substring match;
  others---recall of gold strings.
\item \textbf{Statistics}: per-instance paired bootstrap $B{=}50{,}000$;
  CRC-based seeds (Python \texttt{hash()} is salted per process and
  irreproducible); cross-model flips by joint instance-aligned resampling.
\item \textbf{Compute}: single RTX 3090 48\,GB (rented),
  ${\sim}$US\$15 total; a brief H800 comparison measured only $1.67\times$
  (sdpa) / $1.13\times$ (eager) speedup at $5\times$ the price.
\item \textbf{Data hygiene}: 331 OOM holes detected by coverage assertion
  and refilled; 4{,}932 invalid gemma rows quarantined with checksums; all
  raw JSONL retained.
\end{itemize}

\section{Per-axis results, query-aware arm}
\label{app:aware}

\begin{table}[h]
\caption{Per-axis mean gap vs the \besttriv (12 cells per entry), aware
arm. Compare Table~\ref{tab:axes}. $^\dagger$backend-confounded; no ranking
claim (\S\ref{sec:backend}).}
\label{tab:axes-aware}
\begin{center}
\small
\begin{tabular}{lccc}
\toprule
\textbf{Press} & \textbf{Retrieval} & \textbf{Aggregation} & \textbf{Multi-hop} \\
\midrule
\keydiff          & $+0.262$ (12/12) & $+0.213$ (11/12) & $+0.073$ (10/12) \\
TOVA              & $+0.269$ (12/12) & $+0.174$ (10/12) & $+0.151$ (11/12) \\
AdaKV             & $+0.181$ (8/12)  & $+0.130$ (9/12)  & $+0.152$ (11/12) \\
ExpectedAttention & $-0.033$ (4/12)  & $-0.098$ (2/12)  & $+0.050$ (9/12) \\
\snapkv           & $+0.144$ (8/12)  & $+0.101$ (6/12)  & $+0.152$ (11/12) \\
\midrule
\hho$^\dagger$    & $-0.213$ (0/12)  & $-0.109$ (2/12)  & $-0.013$ (7/12) \\
\bottomrule
\end{tabular}
\end{center}
\end{table}

\end{document}